\documentclass[10pt,twocolumn,letterpaper]{article}

\usepackage{cvpr}
\usepackage{times}
\usepackage{epsfig}
\usepackage{graphicx}
\usepackage{amsmath}
\usepackage{amssymb}
\usepackage{paralist}
\usepackage{multirow}
\usepackage{booktabs}
\usepackage{url}
\usepackage{multirow}
\usepackage{makecell}

\usepackage{amssymb}
\usepackage{accents}
\usepackage{comment}
\usepackage{booktabs}
\usepackage{enumitem}
\usepackage{makecell,multirow,tabularx}
\usepackage{booktabs}
\usepackage{multirow}
\usepackage{colortbl}
\usepackage{times}
 \usepackage{multirow}
 
\usepackage[pagebackref=true,breaklinks=true,letterpaper=true,colorlinks,bookmarks=false]{hyperref}
\usepackage{float}
\usepackage{booktabs}
\usepackage{multirow}
\usepackage{graphicx}
\usepackage[table,xcdraw]{xcolor}

\begin{document}

\title{Unsupervised Fingerphoto Presentation Attack Detection With Diffusion Models}
\author{Hailin Li \quad Raghavendra Ramachandra\\
Norwegian University of Science and Technology\\
Teknologivegen 22, 2815 Gj\o vik, Norway\\
{\tt\small \{hailin.li~|~raghavendra.ramachandra\}@ntnu.no}
\and
Mohamed Ragab \quad
Soumik Mondal \quad
Yong Kiam Tan\quad
Khin Mi Mi Aung \\
Institute for Infocomm Research (I$^2$R),
Agency for Science, Technology and Research (A$^*$STAR)\\
1 Fusionopolis Way, \#21-01, Connexis South Tower, Singapore 138632, Republic of Singapore\\
{\tt\small mohamedr002@e.ntu.edu.sg \quad \{soumikm~|~tanyk1~|~mmaung\}@i2r.a-star.edu.sg}
}

\maketitle
 \thispagestyle{empty}

\begin{abstract}
 Smartphone-based contactless fingerphoto authentication has become a reliable alternative to traditional contact-based fingerprint biometric systems owing to rapid advances in smartphone camera technology.
   Despite its convenience, fingerprint authentication through fingerphotos is more  vulnerable to presentation attacks, which has motivated recent research efforts towards developing fingerphoto Presentation Attack Detection (PAD) techniques. However, prior PAD approaches utilized supervised learning methods that require labeled training data for both bona fide and attack samples.
   This can suffer from two key issues, namely (i) generalization---the detection of novel presentation attack instruments (PAIs) unseen in the training data, and (ii) scalability---the collection of a large dataset of attack samples using different PAIs. To address these challenges, we propose a novel unsupervised approach based on a state-of-the-art deep-learning-based diffusion model, the Denoising Diffusion Probabilistic Model (DDPM), which is trained solely on bona fide samples. The proposed approach detects  Presentation Attacks (PA) by calculating the reconstruction similarity between the input and output pairs of the DDPM.
   We present extensive experiments across three PAI datasets to test the accuracy and generalization capability of our approach.
   The results show that the proposed DDPM-based PAD method achieves significantly better detection error rates on several PAI classes compared to other baseline unsupervised approaches.
\end{abstract}

\section{Introduction}
\label{sec:intro}

The recent prevalence of smartphones has engendered a dual range of consequences for biometric authentication. On the one hand, the ubiquity of smartphones has facilitated the deployment of biometric verification methods such as facial, vocal, and fingerprint recognition systems. On the other hand, biometric authentication, which is often underpinned by machine learning models, is beset with practical security and privacy issues that have emerged in real-world scenarios~\cite{DBLP:conf/sp/ChenCFDZSL21,DBLP:conf/ccs/FredriksonJR15,DBLP:conf/ccs/SharifBBR16}.
This necessitates the adoption of adequate defensive measures when designing and deploying biometric authentication in the real world~\cite{DBLP:journals/tbbis/JainDE22}.

We focus on \emph{fingerphotos}, which are high-quality images of a user's fingertip portion, for example, those captured using a smartphone camera.
Fingerphoto biometrics is a promising technology owing to the wide availability of smartphone cameras, the ability to perform contactless fingerphoto capture, and the lack of requirements for specialized capture devices (unlike traditional fingerprints \cite{maltoni2009handbook}).
Nevertheless, fingerphoto authentication shares the aforementioned security flaws, including vulnerability to \emph{presentation attacks}, where spoof materials with fingerprint-like textures are presented to the camera.

Naturally, the problem of fingerphoto \emph{presentation attack detection} (PAD) has been investigated extensively in prior work~\cite{li2023deep,marasco2022deep,purnapatra2023presentation,taneja2016fingerphoto,  wasnik2018presentation,  zhang20162d}.
However, all of these prior approaches consider PAD in the \emph{supervised} setting, i.e., where labeled training samples are available for both bona fide and spoof fingerphotos.
We observe that
\begin{inparaenum}[\it (i)]
 \item in practice, bona fide samples are much easier to obtain than spoof samples and
 \item models trained only on certain types of presentation attack samples may suffer from the ``unseen materials'' problem~\cite{DBLP:journals/tbbis/JainDE22}, with a lack of generalization to new materials for which there is no pre-existing data.
\end{inparaenum}

\begin{figure}[t]
\newcommand{\fwidth}{0.05}
\centering
\begin{tabular}{@{}ccccc@{}}
\toprule
Input & CAE & VAE & SGA & DDPM \\
\midrule
\includegraphics[width=\fwidth\textwidth]{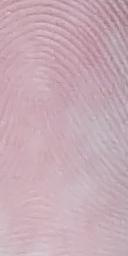} &
\includegraphics[width=\fwidth\textwidth]{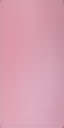} &
\includegraphics[width=\fwidth\textwidth]{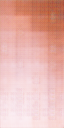} &
\includegraphics[width=\fwidth\textwidth]{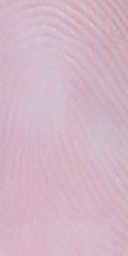} &
\includegraphics[width=\fwidth\textwidth]{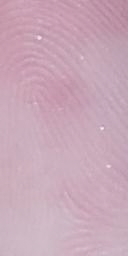}\\ \midrule
\includegraphics[width=\fwidth\textwidth]{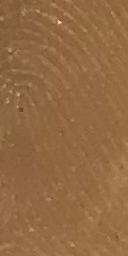} &
\includegraphics[width=\fwidth\textwidth]{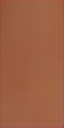} &
\includegraphics[width=\fwidth\textwidth]{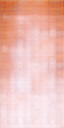} &
\includegraphics[width=\fwidth\textwidth]{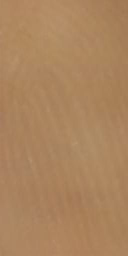} &
\includegraphics[width=\fwidth\textwidth]{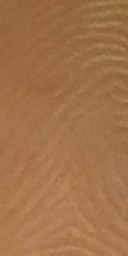}\\\midrule
\includegraphics[width=\fwidth\textwidth]{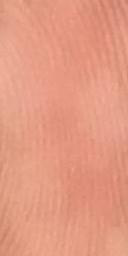} &\includegraphics[width=\fwidth\textwidth]{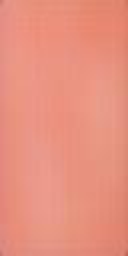} &
\includegraphics[width=\fwidth\textwidth]{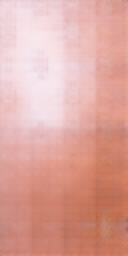} &
\includegraphics[width=\fwidth\textwidth]{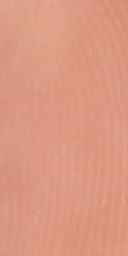} &
\includegraphics[width=\fwidth\textwidth]{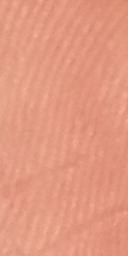}\\\midrule
\includegraphics[width=\fwidth\textwidth]{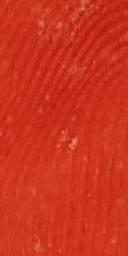} &
\includegraphics[width=\fwidth\textwidth]{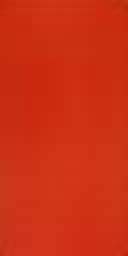} &
\includegraphics[width=\fwidth\textwidth]{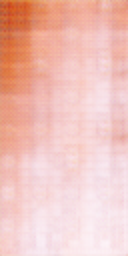} &
\includegraphics[width=\fwidth\textwidth]{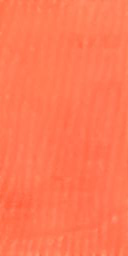} &
\includegraphics[width=\fwidth\textwidth]{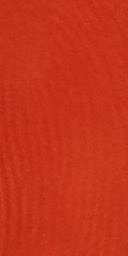}\\\midrule
\includegraphics[width=\fwidth\textwidth]{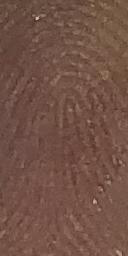} &
\includegraphics[width=\fwidth\textwidth]{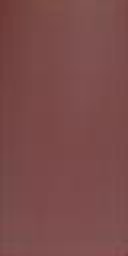} &
\includegraphics[width=\fwidth\textwidth]{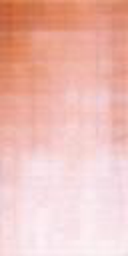} &
\includegraphics[width=\fwidth\textwidth]{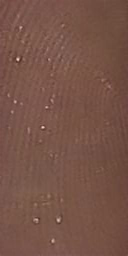} &
\includegraphics[width=\fwidth\textwidth]{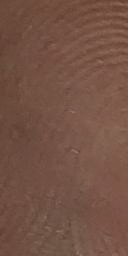}
\end{tabular}
   \caption{Fingerphoto reconstructions using various unsupervised models from an input image (top-to-bottom): bona fide fingerphoto, then PAI attack images using Ecoflex, photopaper, Playdoh, and woodglue; models (left-to-right): convolutional autoencoder (CAE), variational autoencoder (VAE), StyleGAN-ADA (SGA), and DDPM; the input images are not seen during model training.}
\label{fig:fingerphotosamples}
\end{figure}

To tackle these challenges, we formulate fingerphoto presentation attack detection as a \emph{one-class} unsupervised classification problem, which enables us to train PAD models using \emph{only} bona fide fingerphoto training samples.
In related literature on unsupervised classification~\cite{engelsma2019generalizing,kolberg2021anomaly,liu2021one,rohrer2021gan}, an auto-encoder is typically used for image reconstruction.
However, we have empirically observed (cf.~Figure~\ref{fig:fingerphotosamples}) that auto-encoder models offer poor reconstruction quality for bona fide fingerphotos. We hypothesize that using a model which can learn to capture finer details of fingerprints may lead to better PAD performance.
This motivates our search for alternative generative models which can produce high-fidelity fingerphotos with strong PAD performance.

The contributions of this study are as follows:
\begin{itemize}[leftmargin=*,noitemsep, topsep=0pt,parsep=0pt,partopsep=0pt]
\item We propose the use of a deep learing-based, Denoising Diffusion Probabilistic Model (DDPM)~\cite{ho2020denoising} for fingerphoto synthesis; we visually observe (see Figure~\ref{fig:fingerphotosamples}), that state-of-the-art generative models such as DDPM can reproduce highly realistic fingerphoto images despite being trained with a limited dataset.
\item We show how to turn a trained synthesis model into a one-class unsupervised PAD model using the Learned Perceptual Image Patch Similarity (LPIPS) metric~\cite{zhang2018unreasonable}.
\item We experimentally show across three datasets that DDPM equipped with LPIPS outperforms other one-class classifier baselines in detection error rates. 
\item Finally, we conduct extensive experiments with variations of the training and detection process to investigate key factors contributing to our method's performance.
\end{itemize}

The rest of the paper is organized as follows: Section~\ref{sec:related} presents related work on fingerphoto PAD methods.
Section~\ref{sec:method} introduces our proposed method, namely the model architecture and similarity metric;
Section~\ref{sec:experiment} presents a comprehensive suite of experiments for evaluating the proposed method.
Section~\ref{sec:conclusion} concludes with some future directions.

\begin{table*}[htbp]
\centering
\caption{Existing smartphone-based contactless fingerprint/fingerphoto PAD methods.}
\label{tab:SmartphoneFPAD}
\begin{tabular}{|c|c|p{10.355em}|p{13.355em}|p{6em}|}
\hline
 Author& Year & Method & Database and PAIs & Supervised or Unsupervised\\
\hline
 Taneja et al. \cite{taneja2016fingerphoto} & 2016 & Hand-crafted based approach & 1536 bona fide and 4096 spoofed images with  two PAIs & Supervised\\
 \hline
 Zhang et al. \cite{zhang20162d} & 2016 & CNN and hand-crafted based approach & 67011 bona fide and 65581 attack samples with three PAIs& Supervised\\
\hline
Wasnik et al. \cite{wasnik2018presentation} & 2018 &  LBP, BSIF and HOG with SVM & 50 subjects
consisting of three sessions of bona fide data and three PAIs & Supervised\\
\hline
Marasco et al. \cite{marasco2022deep}& 2022 & AlexNet, DenseNet201, DenseNet121, ResNet18, ResNet34, MobileNet-V2  & 4096 genuine and 8192 spoofed images with three PAIs& Supervised\\
\hline
Purnapatra et al. \cite{purnapatra2023presentation}& 2023 & DenseNet 121 and NASNet &  14000 bona fide and 1000 attack samples with five PAIs & Supervised \\
\hline
Li et al. \cite{li2023deep} & 2023 & AlexNet, DenseNet201, MobileNet-V2, NASNet, ResNet50, GoogleNet, EfficientNet-B0  and Vision Transformers &  5886 bona fide and 4247 attack samples with four PAIs & Supervised\\
\hline
Adami et al. \cite{adami2023universal} & 2023 & StyleGAN-ADA and ResNet 18&  5886 bona fide and 4247 attack samples with four PAIs & Supervised\\
\hline
\textbf{This work} &\textbf{2024}  &\textbf{Denoising Diffusion Probabilistic Model, LPIPS for classification} & \textbf{Three datasets of 10886 bona fide and 12035 attack samples with 19 PAIs}  & \textbf{Unsupervised} \\
\hline
\end{tabular}
\end{table*}

\section{Related work}
\label{sec:related}

Fingerprint PAD has been studied for over a decade~\cite{chugh2019fingerprint,marcialis2009first,park2019presentation}, whereas fingerphoto PAD is a more recent topic. Publicly available datasets such as those from Purnaputra et al.~\cite{purnapatra2023presentation} and Kolberg et al.~\cite{kolberg2023colfispoof} have helped to drive and enable research in this crucial latter area.

An overview of prior methods for fingerphoto PAD is given in Table~\ref{tab:SmartphoneFPAD} (adapting an earlier table by Li et al.~\cite{li2023deep}).
Earlier approaches utilized handcrafted features extracted from images together with classical supervised learning models, such as support vector machines (SVM)~\cite{taneja2016fingerphoto,wasnik2018presentation}.
Zhang et al.~\cite{zhang20162d} proposed a hybrid 2D fake fingerprint detection based on convolutional neural networks (CNNs) and two local descriptors (Local Binary Pattern and Local Phase Quantization).
More recent approaches rely entirely on deep learning-based feature extraction;
Puranpatra et al.~\cite{purnapatra2023presentation} utilized a combination of two CNN models, DenseNet 121 and NASNet, and evaluated on five different PAIs.
Li et al.~\cite{li2023deep} performed a comparative study utilizing eight different deep models for feature extraction and trained the resulting features with a support vector machine (SVM) targeting unseen attacks;
Adami et al.~\cite{adami2023universal} proposed a semi-supervised learning model which trained a ResNet-18 model with a combination of the Arcface and Center Loss functions using live samples and synthetic spoofed samples generated by StyleGAN-ADA~\cite{karras2020training}.  Our comparative novelty is the combined use of a diffusion model and LPIPS to achieve state-of-the-art results for fingerphoto PAD.


\section{Proposed Method}
\label{sec:method}

\begin{figure*}[htp]
\begin{center}
\includegraphics[width=0.85\linewidth]{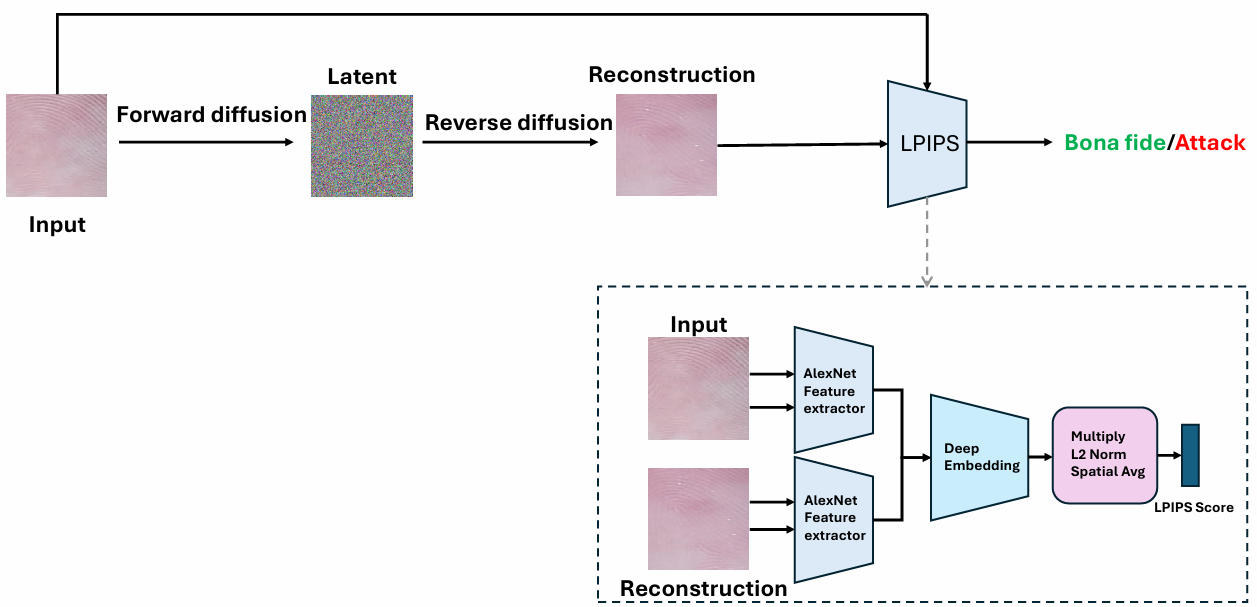}
\end{center}
   \caption{The pipeline of our proposed DDPM and LPIPS fingerphoto PAD method.}
\label{Fig:Pipeline}
\end{figure*}

Figure~\ref{Fig:Pipeline} shows the block diagram of the proposed fingerphoto PAD algorithm based on Denoising Diffusion Probabilistic Model (DDPM) as the generative model combined with Learned Perceptual Image Patch Similarity (LPIPS) as the image similarity metric. 
At a high level, the proposed method consists of two steps:
\begin{enumerate} [leftmargin=*,noitemsep, topsep=0pt,parsep=0pt,partopsep=0pt]
\item First, we train an unsupervised \emph{generative model} to reconstruct fingerphoto Region Of Interest (ROI) images. The ROI is extracted by cropping a $128 \times 256$ region close to the center point. This training process is conducted using only bona fide fingerphotos.
\item Then, to carry out fingerphoto PAD, we apply the generative model directly on the input test image (either bona fide or attack). We expect the model's reconstruction process to work well on bona fide samples, but to perform poorly on attack samples. Accordingly, we calculate the similarity between the input and reconstructed images using an \emph{image similarity metric}. Reconstructions with similarity scores below a predefined threshold are classified as attack samples.
\end{enumerate}


\subsection{Denoising Diffusion Probabilistic Model}
\label{subsec:ddpm}

Diffusion generative models are composed of two opposite processes: forward and reverse diffusion~\cite{ho2020denoising}.
Given a data point $x_{0}$ from the real data distribution $q(x_{0})$, the forward diffusion process gradually destroys its data structure by adding noise.
Specifically, Gaussian noise with variance $\beta_{t}$ is added to $x_{t-1}$ at each step of the Markov chain, producing a new latent variable $x_{t}$ with distribution $q(x_{t} \mid x_{t-1})$. This process is formulated as follows: 
\begin{equation}
    q(x_{t} \mid x_{t-1}) = \mathcal{N}(x_{t}; \sqrt{1-\beta_{t}}x_{t-1},\beta_{t} \textbf{I})
    \label{eq:forward1}
\end{equation}
where $\beta_{t}$ is a pre-defined or learned noise variance schedule, and \textbf{I} is the identity matrix. Thus to produce a sample $x_{t}$ the following distribution can be used:
\begin{equation}
    q(x_{0} \mid t_{0}) = \mathcal{N}(x_{t}; \sqrt{\alpha_{t}} x_{0},(1-\alpha_{t}) \textbf{I})
    \label{eq:forward2}
\end{equation}

The reverse diffusion process aims to learn a transition kernel from $x_{t}$ to $x_{t-1}$, which is defined as the following Gaussian distribution:
\begin{equation}
    p_{\theta}(x_{t} \mid x_{t-1}) =  \mathcal{N}(x_{t-1}; \mu_{\theta}(x_{t,t}),\sigma_{t}^2\textbf{I})
    \label{eq:reverse1}
\end{equation}

The data distribution $p(x_{0})$ can be formulated as:
\begin{equation}
    p_{\theta}(x_{0}) = \int p(x_{t}) \prod_{t=1}^{T}  p_{\theta}(x_{t-1} \mid x_{t})dx_{1:T}
    \label{eq:reverse2}
\end{equation}

The model architecture that we used is based on a variation version of the original DDPM model, named DifFace, proposed by Yue~\cite{yue2022difface}. 
This model is designed to recover a high quality (HQ) image from its low-quality (LQ) counterpart so that we can utilize it to enhance the input fingerphoto images.
The motivation behind the DifFace is to replace $p(x_{N} |y_{0})$ with the marginal distribution $q(x_{N} |x_{0})$ defined in Eq \ref{eq:forward2}. 
With the aid of this transition, the posterior distribution $p(x_{0}|y_{0})$ can be constructed as follows:
\begin{equation}
    p(x_{0} |y_{0}) = \int p(x_{N} |y_{0} \prod_{t=1}^{N} p_{\theta}(x_{t-1} \mid x_{t})dx_{1:N}
\end{equation}
 where $1 \le N < T$ is an arbitrary timestep so that $x_{0}$ can be restored from $y_{0}$  from this posterior using ancestral sampling \cite{bishop2006pattern}.
 The posterior distribution $ p (x_0 | y_0) $ aims to infer the HQ image $ x_0$ conditioned on its LQ counterpart $ y_0$.


\subsection{Learned Perceptual Image Patch Similarity}
\label{subsec:similaritymetric}
The Learned Perceptual Image Patch Similarity (LPIPS)~\cite{zhang2018unreasonable} metric calculates perceptual similarity between two images using the similarity of activations for a pre-trained feature extractor, usually a deep learning-based convolutional neural network.
In our experiments, we utilize the LPIPS metric with a pre-trained AlexNet~\cite{krizhevsky2012imagenet}, which was chosen based on its robustness and accuracy.

\section{Experimental Evaluation}
\label{sec:experiment}
In this section, we present an extensive quantitative evaluation of the proposed method by using three different fingerphoto PAD datasets.
In the following sections, we describe the experimental dataset, evaluation metrics, performance evaluation protocol, and performance comparison with six alternative unsupervised fingerphoto PAD methods.

\subsection{Datasets}
In this study, we performed experiments using three datasets: CLARKSON~\cite{purnapatra2023presentation}, NTNU dataset collected by our group \cite{wasnik2018improved}, and the HDA dataset~\cite{kolberg2023colfispoof}.

\begin{figure}[htp]
\begin{center}
\includegraphics[width=0.95\linewidth]{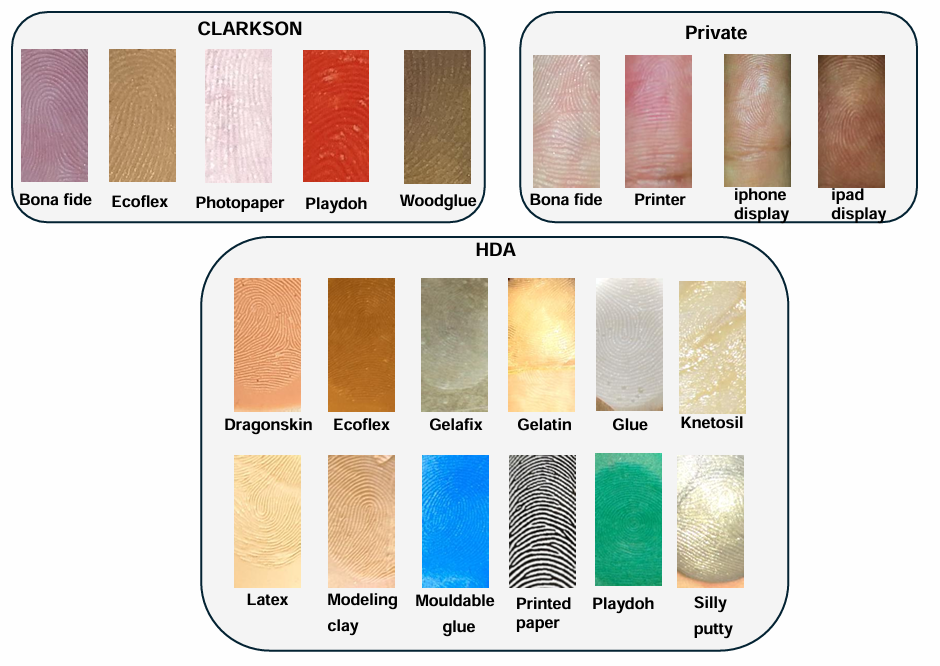}
\end{center}
   \caption{PAIs and bona fide samples from the three datasets.}
\label{Fig:PAI_example}
\end{figure}

The statistics for the CLARKSON~\cite{purnapatra2023presentation} dataset are listed in Table \ref{tab:D1 statistics} which includes three different smartphones for image capture and four different PAIs.
Table \ref{tab:D2 statistics} shows the statistics of the NTNU fingerphoto PAD dataset captured using an Apple iPhone 6s / iPad Pro, which comprises three PAIs.
Finally, we include another testing dataset developed by Kolberg et al.~\cite{kolberg2023colfispoof}.
This dataset contains only attack samples and 12 different PAIs for a total of 7200 samples.
Sample PAI images for each dataset are shown in Figure~\ref{Fig:PAI_example}.

\begin{table}[h]
\centering
\caption{Presentation Attack Instruments (PAIs) statistics, e.g., the number of samples and capture devices, for CLARKSON~\cite{purnapatra2023presentation}. }
\label{tab:D1 statistics}
{\scriptsize\begin{tabular}{|c|c|c|c|c|}
\hline
\multirow{3}{*}{Image type}& \multicolumn{4}{c|}{Number of samples} \\
\cline{2-5}
                  &   \multirow{2}{*}{iPhone 7}    & \multirow{2}{*}{iPhone X} &   Samsung  & \multirow{2}{*}{Total}\\
        & & &Galaxy S9  & \\
        
\hline
Bona fide                  &  858         &  691     & 4336 & 5886 \\
\hline                  
PAI: Ecoflex                  &     832     &  0 & 416& 1248 \\
\hline 
PAI: Photopaper                  &     832      &  272 & 0 &1104\\
\hline
PAI: Playdoh                  &     0      & 0   &1623&1623 \\
\hline
PAI: Woodglue                  &   0 &  272     &  0 &272 \\
\hline
\end{tabular}}
\end{table}
\begin{table}[h]
\caption{Presentation Attack Instruments (PAIs) statistics, e.g., the number of samples and capture devices, for the NTNU dataset. }
\label{tab:D2 statistics}
\centering
{\scriptsize\begin{tabular}{|c|c|c|c|c|}
\hline
\multirow{2}{*}{Image type}& \multicolumn{4}{c|}{Number of samples} \\
\cline{2-5}
                  &   Printer    &  iPhone 6s &   iPad Pro  & Total\\
\hline
Bona fide                  &      0     &   5000    & 0 & 5000 \\
\hline                  
PAI: Printer attack                  &     196     &  0 & 0& 196 \\
\hline 
PAI: iPhone 6s display attack                &     0      &  196 & 0 &196\\
\hline
PAI: iPad Pro display attack               &     0      & 0   &196&196 \\
\hline
\end{tabular}
}
\end{table}

\subsection{Evaluation Metric}
The experimental results were obtained using the standard ISO/IEC 30107-3~\cite{ISO-IEC-30107-3-PAD-metrics-170227} methodology for evaluating biometric systems.
The Attack Presentation Classification Error Rate (APCER) is the percentage ratio of presentation attack test samples misidentified as bona fide samples.
The Bona fide Presentation Classification Error Rate (BPCER) is the percentage ratio of bona fide test samples misidentified as presentation attack examples.

\subsection{Performance Evaluation Protocol}
\label{subsec:methodology}

In this section, we discuss the performance evaluation protocols employed to benchmark the performance of the proposed method and alternative baselines for unsupervised fingerphoto PAD.
We provide precise details of the implemented experimental procedures to enable reproducibility of our results on the datasets.

\paragraph{ROI Extraction.} For all images (training and testing), region-of-interest (ROI) extraction was first applied to detect the main finger portion of the fingerphoto. Li and Ramachandra~\cite{Li_2024_WACV} reported that ROI extraction affects detection performance because the presentation attack samples, e.g., for CLARKSON, may cover only a portion of the finger. Performing ROI extraction on the entire dataset provides a fairer comparison of fingerphoto PAD capability, as it forces the models to detect anomalies in the fingerprint texture, rather than unrelated background information.

\paragraph{Baseline Unsupervised Models.} We evaluate our approach (DDPM) against six different baseline unsupervised algorithms which are labeled in all tables as follows: 
\begin{itemize} [leftmargin=*,noitemsep, topsep=0pt,parsep=0pt,partopsep=0pt]
\item (RN OC SVM) Features of the input bona fide fingerphoto training images are extracted using a ResNet 50~\cite{resnet} pretrained deep feature extractor. These extracted features are then used to train a one-class SVM (OC SVM) for anomaly (i.e., out-of-class) detection.
\item (DN OC SVM) Similar to the above, except the feature extraction is done using a DenseNet 121~\cite{densenet} pretrained model instead.
\item (ViT OC SVM) Similar to the above, except the feature extraction is done using a vision transformer model~\cite{ViT}.
\item (CAE) The convolutional auto-encoder model~\cite{zhang2018better}.
\item (VAE) The variational auto-encoder model~\cite{kingma2013auto}.
\item (SGA) We used the GAN variation known as StyleGAN-ADA~\cite{karras2020training}, which proposes an adaptive discriminator augmentation mechanism that significantly stabilizes training in limited data regimes (such as for our smaller fingerphoto training datasets).

\end{itemize}

\paragraph{Dataset Partition.} The datasets are partitioned for training and testing as shown in Table~\ref{tab:datasetpartition}.

\begin{table}[h]
\centering
\caption{Summary of dataset partition statistics (``BF'' is for bona fide and ``A'' is for attack samples).}
\label{tab:datasetpartition}
{\small\begin{tabular}{|l|l|l|l|}
\hline
\textbf{Dataset}          & \textbf{Type}     & \textbf{Samples} & \textbf{Subjects} \\ \hline
\multirow{3}{*}{CLARKSON~\cite{purnapatra2023presentation}} & BF (train) & 4656             & 21                \\ \cline{2-4} 
                          & BF (test)  & 1230             & 5                 \\ \cline{2-4} 
                          & A (test)     & All (4 PAIs)     & -                 \\ \hline
\multirow{3}{*}{NTNU~\cite{wasnik2018improved}}  & BF (train) & 4000             & 160               \\ \cline{2-4} 
                          & BF (test)  & 1000             & 40                \\ \cline{2-4} 
                          & A (test)     & All (3 PAIs)     & -                 \\ \hline
HDA~\cite{kolberg2023colfispoof}                       & A (test)     & All (12 PAIs)    & -                 \\ \hline
\end{tabular}}
\end{table}

We use all attack samples from all datasets for testing because our experiments use unsupervised models that do not need attack samples for training
In contrast, we need to partition the bona fide samples into training and testing partitions, as shown in the table. The training and testing partitions have disjoint data subjects.

\paragraph{Experiments.} We run three experiments with different training and test datasets which allows us to evaluate the robustness of fingerphoto PAD methods on different types of PAI, capture devices, and environmental conditions.
The detailed experimental setup is listed in Table~\ref{tab:experimentalsetup}.

\begin{table}[h]
\centering
\caption{Summary of experimental setup.}
\label{tab:experimentalsetup}
{\small\begin{tabular}{|l|l|l|l|}
\hline
\textbf{Expt.} & \textbf{\begin{tabular}[c]{@{}l@{}}Training Set\\ (Bona fide)\end{tabular}}               & \textbf{\begin{tabular}[c]{@{}l@{}}Test Set\\ (Bona fide)\end{tabular}}                   & \textbf{\begin{tabular}[c]{@{}l@{}}Test Set\\ (Attack)\end{tabular}} \\ \hline
\multirow{3}{*}{1} & \multirow{3}{*}{CLARKSON}                                                                 & CLARKSON                                                                                  & CLARKSON                                                    \\ \cline{3-4} 
                   &                                                                                           & NTNU                                                                                   & NTNU                                                     \\ \cline{3-4} 
                   &                                                                                           & CLARKSON                                                                                  & HDA                                                         \\ \hline
\multirow{3}{*}{2} & \multirow{3}{*}{NTNU}                                                                  & CLARKSON                                                                                  & CLARKSON                                                    \\ \cline{3-4} 
                   &                                                                                           & NTNU                                                                                   & NTNU                                                     \\ \cline{3-4} 
                   &                                                                                           & NTNU                                                                                   & HDA                                                         \\ \hline
\multirow{3}{*}{3} & \multirow{3}{*}{\begin{tabular}[c]{@{}l@{}}Combined\\ CLARKSON\\ \& NTNU\end{tabular}} & \multirow{3}{*}{\begin{tabular}[c]{@{}l@{}}Combined\\ CLARKSON\\ \& NTNU\end{tabular}} & CLARKSON                                                    \\ \cline{4-4} 
                   &                                                                                           &                                                                                           & NTNU                                                     \\ \cline{4-4} 
                   &                                                                                           &                                                                                           & HDA                                                         \\ \hline
\end{tabular}}
\end{table}




Our choice of the dataset pairs for each experiment ensures that the results are unbiased.
Since the HDA dataset does not come with bona fide samples, we test it against the bona fide test set of the respective training sets.

\begin{table*}[t]
\centering
\caption{BPCER for various unsupervised PAD methods against PAIs for Experiment 1 (best method in \textbf{bold} for each PAI).}
\label{tab:EER compare table experiment1}
{\small
\begin{tabular}{|c|c|ccccccc|}
\hline
\multirow{3}{*}{Testing dataset} & \multirow{3}{*}{PAI} & \multicolumn{7}{c|}{Unsupervised fingerphoto presentation attack detection}  \\ 
\cline{3-9}   &       & \multicolumn{7}{c|}{BPCER @ APCER = 10(\%)}    \\ 
\cline{3-9}   &       & \multicolumn{1}{c|}{RN OC SVM} & \multicolumn{1}{c|}{DN OC SVM} & \multicolumn{1}{c|}{ViT OC SVM} & \multicolumn{1}{c|}{CAE} & \multicolumn{1}{c|}{VAE} & \multicolumn{1}{c|}{SGA} &  \textbf{DDPM(Ours)}\\ 
\hline
\multicolumn{1}{|c|}{\multirow{4}{*}{CLARKSON}} &     Ecoflex   & \multicolumn{1}{c|}{89.37} & \multicolumn{1}{c|}{70.28} & \multicolumn{1}{c|}{39.76} & \multicolumn{1}{c|}{45.03} & \multicolumn{1}{c|}{62.14} & \multicolumn{1}{c|}{59.99} &  \textbf{6.34}
\\ \cline{2-9}    &   Photopaper                & \multicolumn{1}{c|}{89.61} & \multicolumn{1}{c|}{75.44} & \multicolumn{1}{c|}{58.92} & \multicolumn{1}{c|}{95.32} & \multicolumn{1}{c|}{92.80} & \multicolumn{1}{c|}{93.31} &  \textbf{12.89}\\ 
\cline{2-9}    &     Playdoh    & \multicolumn{1}{c|}{92.07} & \multicolumn{1}{c|}{31.87} & \multicolumn{1}{c|}{\textbf{6.30}} & \multicolumn{1}{c|}{36.12} & \multicolumn{1}{c|}{11.26} & \multicolumn{1}{c|}{81.64} & 9.46\\ 
\cline{2-9}    &     Woodglue              & \multicolumn{1}{c|}{83.58} & \multicolumn{1}{c|}{26.95} & \multicolumn{1}{c|}{32.11} & \multicolumn{1}{c|}{12.37} & \multicolumn{1}{c|}{23.17} & \multicolumn{1}{c|}{37.13} &  \textbf{3.64}\\ \hline

\hline 
\hline 
\multicolumn{1}{|c|}{\multirow{3}{*}{NTNU}} & Paper printout   & \multicolumn{1}{c|}{98.10} & \multicolumn{1}{c|}{93.50} & \multicolumn{1}{c|}{89.80} & \multicolumn{1}{c|}{98.50} & \multicolumn{1}{c|}{97.50} & \multicolumn{1}{c|}{95.40} & \textbf{88.10} \\ 
\cline{2-9}   &   IPhone display   & \multicolumn{1}{c|}{97.80} & \multicolumn{1}{c|}{95.40} & \multicolumn{1}{c|}{83.20} & \multicolumn{1}{c|}{98.30} & \multicolumn{1}{c|}{86.60} & \multicolumn{1}{c|}{85.10} & \textbf{83.00}\\ 
\cline{2-9}   &  iPad display  & \multicolumn{1}{c|}{92.30} & \multicolumn{1}{c|}{90.90} & \multicolumn{1}{c|}{\textbf{76.00}} & \multicolumn{1}{c|}{99.80} & \multicolumn{1}{l|}{86.00} & \multicolumn{1}{c|}{80.40} & 95.90 \\ \hline

\hline 
\hline
\multicolumn{1}{|c|}{\multirow{13}{*}{HDA}} &  Dragonskin & \multicolumn{1}{c|}{88.74} & \multicolumn{1}{c|}{92.07} & \multicolumn{1}{c|}{47.45} & \multicolumn{1}{c|}{86.14} & \multicolumn{1}{c|}{71.32} & \multicolumn{1}{c|}{76.84} & \textbf{8.24}\\ 
\cline{2-9}  &    Ecoflex    & \multicolumn{1}{c|}{83.86} & \multicolumn{1}{c|}{90.41} & \multicolumn{1}{c|}{36.61} & \multicolumn{1}{c|}{80.57} & \multicolumn{1}{c|}{73.24} & \multicolumn{1}{c|}{83.31} & \textbf{5.51}\\ 
\cline{2-9} &  Gelafix     & \multicolumn{1}{c|}{85.56} & \multicolumn{1}{c|}{88.88} & \multicolumn{1}{c|}{51.51} & \multicolumn{1}{l|}{56.22} & \multicolumn{1}{c|}{57.36} & \multicolumn{1}{c|}{62.54} & \textbf{6.75} \\ 
\cline{2-9}   &  Gelatin  & \multicolumn{1}{c|}{79.01} & \multicolumn{1}{c|}{81.26} & \multicolumn{1}{c|}{33.84} & \multicolumn{1}{c|}{31.73} & \multicolumn{1}{c|}{29.49} & \multicolumn{1}{c|}{34.80} &\textbf{7.14}  \\ 
 \cline{2-9}  & Glue   & \multicolumn{1}{c|}{87.53} & \multicolumn{1}{c|}{92.86} & \multicolumn{1}{c|}{60.44} & \multicolumn{1}{c|}{20.33} & \multicolumn{1}{c|}{71.24} & \multicolumn{1}{c|}{86.72} &\textbf{9.66}  \\ 
\cline{2-9} &   Knetosil     & \multicolumn{1}{c|}{81.78} & \multicolumn{1}{c|}{86.49} & \multicolumn{1}{c|}{37.27} & \multicolumn{1}{c|}{91.10} & \multicolumn{1}{c|}{28.66} & \multicolumn{1}{c|}{72.39} & \textbf{3.43} \\ 
\cline{2-9}  &  Latex    & \multicolumn{1}{c|}{87.53} & \multicolumn{1}{c|}{92.66} & \multicolumn{1}{c|}{48.60} & \multicolumn{1}{c|}{84.38} & \multicolumn{1}{c|}{81.11} & \multicolumn{1}{c|}{70.44} & \textbf{3.29}\\
\cline{2-9} &   Modelling clay   & \multicolumn{1}{c|}{88.40} & \multicolumn{1}{c|}{82.37} & \multicolumn{1}{c|}{57.05} & \multicolumn{1}{c|}{49.19} & \multicolumn{1}{c|}{57.85} & \multicolumn{1}{c|}{61.20} &\textbf{0.83}  \\
\cline{2-9}    &  Mouldable glue  & \multicolumn{1}{c|}{87.95} & \multicolumn{1}{c|}{89.16} & \multicolumn{1}{c|}{45.65} & \multicolumn{1}{c|}{90.89} & \multicolumn{1}{c|}{96.21} & \multicolumn{1}{c|}{93.72} & \textbf{2.18}\\ 
 \cline{2-9}  &  Paper printout  & \multicolumn{1}{c|}{85.21} & \multicolumn{1}{c|}{74.99} & \multicolumn{1}{c|}{46.03} & \multicolumn{1}{c|}{8.66} & \multicolumn{1}{c|}{33.12} & \multicolumn{1}{c|}{18.55} & \textbf{0}\\ 
\cline{2-9} & Playdoh    & \multicolumn{1}{c|}{87.11} & \multicolumn{1}{c|}{84.45} & \multicolumn{1}{c|}{44.68} & \multicolumn{1}{c|}{72.05} & \multicolumn{1}{c|}{79.95} & \multicolumn{1}{c|}{87.13} & \textbf{6.75}\\ 
\cline{2-9}  &  Silly putty   & \multicolumn{1}{c|}{89.40} & \multicolumn{1}{c|}{79.11} & \multicolumn{1}{c|}{44.34} & \multicolumn{1}{c|}{88.57} & \multicolumn{1}{c|}{29.95} & \multicolumn{1}{c|}{52.44} & \textbf{7.27}\\
 \hline
\end{tabular}}
\end{table*}

\subsection{Result and Discussion}

\begin{table*}[]
\centering
\caption{BPCER for various unsupervised PAD methods against PAIs for Experiment 2 (best method in \textbf{bold} for each PAI).}
\label{tab:EER compare table experiment2}
{\small
\begin{tabular}{|c|c|ccccccc|}
\hline
\multirow{3}{*}{Testing dataset} & \multirow{3}{*}{PAI} & \multicolumn{7}{c|}{Unsupervised fingerphoto presentation attack detection}  \\ 
\cline{3-9}   &       & \multicolumn{7}{c|}{BPCER @ APCER = 10(\%)}    \\ 
\cline{3-9}   &       & \multicolumn{1}{c|}{RN OC SVM} & \multicolumn{1}{c|}{DN OC SVM} & \multicolumn{1}{c|}{ViT OC SVM} & \multicolumn{1}{c|}{CAE} & \multicolumn{1}{c|}{VAE} & \multicolumn{1}{c|}{SGA} &  \textbf{DDPM(Ours)}\\ 
\hline
\multicolumn{1}{|c|}{\multirow{4}{*}{CLARKSON}} &     Ecoflex   & \multicolumn{1}{c|}{94.25} & \multicolumn{1}{c|}{93.71} & \multicolumn{1}{c|}{82.40} & \multicolumn{1}{c|}{55.44} & \multicolumn{1}{c|}{61.17} & \multicolumn{1}{c|}{59.98} &  \textbf{23.04}
\\ \cline{2-9}    &   Photopaper                & \multicolumn{1}{c|}{96.50} & \multicolumn{1}{c|}{91.33} & \multicolumn{1}{c|}{88.84} & \multicolumn{1}{c|}{90.60} & \multicolumn{1}{c|}{88.49} & \multicolumn{1}{c|}{93.39} &  \textbf{46.33}\\ 
\cline{2-9}    &     Playdoh    & \multicolumn{1}{c|}{87.61} & \multicolumn{1}{c|}{77.23} & \multicolumn{1}{c|}{80.20} & \multicolumn{1}{c|}{73.34} & \multicolumn{1}{c|}{67.52} & \multicolumn{1}{c|}{70.31} & \textbf{28.49}\\ 
\cline{2-9}    &     Woodglue              & \multicolumn{1}{c|}{96.11} & \multicolumn{1}{c|}{88.74} & \multicolumn{1}{c|}{66.52} & \multicolumn{1}{c|}{54.10} & \multicolumn{1}{c|}{62.35} & \multicolumn{1}{c|}{60.13} &  \textbf{12.45}\\ \hline

\hline 
\hline 
\multicolumn{1}{|c|}{\multirow{3}{*}{NTNU}} & Paper printout   & \multicolumn{1}{c|}{94.60} & \multicolumn{1}{c|}{95.60} & \multicolumn{1}{c|}{90.30} & \multicolumn{1}{c|}{95.70} & \multicolumn{1}{c|}{96.10} & \multicolumn{1}{c|}{94.50} & \textbf{34.20} \\ 
\cline{2-9}   &   IPhone display   & \multicolumn{1}{c|}{95.20} & \multicolumn{1}{c|}{94.00} & \multicolumn{1}{c|}{92.70} & \multicolumn{1}{c|}{88.60} & \multicolumn{1}{c|}{85.90} & \multicolumn{1}{c|}{88.10} & \textbf{35.50}\\ 
\cline{2-9}   &  iPad display  & \multicolumn{1}{c|}{97.70} & \multicolumn{1}{c|}{93.90} & \multicolumn{1}{c|}{91.70} & \multicolumn{1}{c|}{94.40} & \multicolumn{1}{c|}{87.90} & \multicolumn{1}{c|}{85.60} & \textbf{39.90} \\ \hline

\hline 
\hline
\multicolumn{1}{|c|}{\multirow{13}{*}{HDA}} &  Dragonskin & \multicolumn{1}{c|}{93.20} & \multicolumn{1}{c|}{69.10} & \multicolumn{1}{c|}{71.70} & \multicolumn{1}{c|}{91.00} & \multicolumn{1}{c|}{\textbf{68.90}} & \multicolumn{1}{c|}{69.10} & 90.70\\ 
\cline{2-9}  &    Ecoflex    & \multicolumn{1}{c|}{89.80} & \multicolumn{1}{c|}{94.70} & \multicolumn{1}{c|}{54.70} & \multicolumn{1}{c|}{53.80} & \multicolumn{1}{c|}{61.10} & \multicolumn{1}{c|}{63.30} & \textbf{52.10}\\ 
\cline{2-9} &  Gelafix     & \multicolumn{1}{c|}{72.80} & \multicolumn{1}{c|}{94.60} & \multicolumn{1}{c|}{76.00} & \multicolumn{1}{l|}{86.50} & \multicolumn{1}{c|}{67.50} & \multicolumn{1}{c|}{\textbf{62.40}} & 68.20 \\ 
\cline{2-9}   &  Gelatin  & \multicolumn{1}{c|}{78.00} & \multicolumn{1}{c|}{97.60} & \multicolumn{1}{c|}{70.30} & \multicolumn{1}{c|}{38.50} & \multicolumn{1}{c|}{41.00} & \multicolumn{1}{c|}{48.90} &\textbf{37.60}  \\ 
 \cline{2-9}  & Glue   & \multicolumn{1}{c|}{78.60} & \multicolumn{1}{c|}{75.90} & \multicolumn{1}{c|}{78.50} & \multicolumn{1}{c|}{54.70} & \multicolumn{1}{c|}{61.40} & \multicolumn{1}{c|}{63.30} &\textbf{50.10}  \\ 
\cline{2-9} &   Knetosil     & \multicolumn{1}{c|}{83.30} & \multicolumn{1}{c|}{95.30} & \multicolumn{1}{c|}{59.60} & \multicolumn{1}{c|}{26.80} & \multicolumn{1}{c|}{30.90} & \multicolumn{1}{c|}{31.50} & \textbf{24.80} \\ 
\cline{2-9}  &  Latex    & \multicolumn{1}{c|}{96.20} & \multicolumn{1}{c|}{52.20} & \multicolumn{1}{c|}{79.30} & \multicolumn{1}{c|}{24.10} & \multicolumn{1}{c|}{49.50} & \multicolumn{1}{c|}{45.20} & \textbf{22.90}\\
\cline{2-9} &   Modelling clay   & \multicolumn{1}{c|}{87.80} & \multicolumn{1}{c|}{77.40} & \multicolumn{1}{c|}{81.60} & \multicolumn{1}{c|}{\textbf{24.30}} & \multicolumn{1}{c|}{29.60} & \multicolumn{1}{c|}{34.60} &26.20  \\
\cline{2-9}    &  Mouldable glue  & \multicolumn{1}{c|}{85.20} & \multicolumn{1}{c|}{82.50} & \multicolumn{1}{c|}{70.80} & \multicolumn{1}{c|}{23.20} & \multicolumn{1}{c|}{27.80} & \multicolumn{1}{c|}{41.10} & \textbf{21.70}\\ 
 \cline{2-9}  &  Paper printout  & \multicolumn{1}{c|}{83.00} & \multicolumn{1}{c|}{77.90} & \multicolumn{1}{c|}{62.50} & \multicolumn{1}{c|}{8.10} & \multicolumn{1}{c|}{35.50} & \multicolumn{1}{c|}{43.90} & \textbf{0.10}\\ 
\cline{2-9} & Playdoh    & \multicolumn{1}{c|}{88.00} & \multicolumn{1}{c|}{91.00} & \multicolumn{1}{c|}{68.90} & \multicolumn{1}{c|}{59.20} & \multicolumn{1}{c|}{\textbf{44.70}} & \multicolumn{1}{c|}{70.30} & 68.70\\ 
\cline{2-9}  &  Silly putty   & \multicolumn{1}{c|}{81.40} & \multicolumn{1}{c|}{90.90} & \multicolumn{1}{c|}{75.80} & \multicolumn{1}{c|}{57.80} & \multicolumn{1}{c|}{55.50} & \multicolumn{1}{c|}{61.00} & \textbf{48.30}\\
 \hline
\end{tabular}}
\end{table*}

\begin{table*}[]
\centering
\caption{BPCER for various unsupervised PAD methods against PAIs for Experiment 3 (best method in \textbf{bold} for each PAI).}
\label{tab:EER compare table experiment3}
{\small
\begin{tabular}{|c|c|ccccccc|}
\hline
\multirow{3}{*}{Testing dataset} & \multirow{3}{*}{PAI} & \multicolumn{7}{c|}{Unsupervised fingerphoto presentation attack detection}  \\ 
\cline{3-9}   &       & \multicolumn{7}{c|}{BPCER @ APCER = 10(\%)}    \\ 
\cline{3-9}   &       & \multicolumn{1}{c|}{RN OC SVM} & \multicolumn{1}{c|}{DN OC SVM} & \multicolumn{1}{c|}{ViT OC SVM} & \multicolumn{1}{c|}{CAE} & \multicolumn{1}{c|}{VAE} & \multicolumn{1}{c|}{SGA} &  \textbf{DDPM(Ours)}\\ 
\hline
\multicolumn{1}{|c|}{\multirow{4}{*}{CLARKSON}} &     Ecoflex   & \multicolumn{1}{c|}{88.87} & \multicolumn{1}{c|}{94.16} & \multicolumn{1}{c|}{66.24} & \multicolumn{1}{c|}{47.17} & \multicolumn{1}{c|}{42.33} & \multicolumn{1}{c|}{45.64} &  \textbf{21.07}
\\ \cline{2-9}    &   Photopaper                & \multicolumn{1}{c|}{89.10} & \multicolumn{1}{c|}{95.70} & \multicolumn{1}{c|}{83.07} & \multicolumn{1}{c|}{93.70} & \multicolumn{1}{c|}{93.51} & \multicolumn{1}{c|}{95.53} &  \textbf{40.21}\\ 
\cline{2-9}    &     Playdoh    & \multicolumn{1}{c|}{85.32} & \multicolumn{1}{c|}{64.00} & \multicolumn{1}{c|}{33.14} & \multicolumn{1}{c|}{31.62} & \multicolumn{1}{c|}{32.53} & \multicolumn{1}{c|}{35.69} &  \textbf{20.11} \\ 
\cline{2-9}    &     Woodglue              & \multicolumn{1}{c|}{90.72} & \multicolumn{1}{c|}{86.17} & \multicolumn{1}{c|}{54.40} & \multicolumn{1}{c|}{49.16} & \multicolumn{1}{c|}{53.67} & \multicolumn{1}{c|}{48.97} &  \textbf{12.46}\\ \hline

\hline 
\hline 
\multicolumn{1}{|c|}{\multirow{3}{*}{NTNU}} & Paper printout   & \multicolumn{1}{c|}{85.97} & \multicolumn{1}{c|}{96.53} & \multicolumn{1}{c|}{72.59} & \multicolumn{1}{c|}{90.32} & \multicolumn{1}{c|}{68.54} & \multicolumn{1}{c|}{77.14} & \textbf{24.67} \\ 
\cline{2-9}   &   IPhone display   & \multicolumn{1}{c|}{85.36} & \multicolumn{1}{c|}{90.56} & \multicolumn{1}{c|}{80.18} & \multicolumn{1}{c|}{82.27} & \multicolumn{1}{c|}{74.81} & \multicolumn{1}{c|}{76.49} & \textbf{24.82}\\ 
\cline{2-9}   &  iPad display  & \multicolumn{1}{c|}{89.18} & \multicolumn{1}{c|}{95.47} & \multicolumn{1}{c|}{64.74} & \multicolumn{1}{c|}{65.83} & \multicolumn{1}{l|}{63.20} & \multicolumn{1}{c|}{70.06} & \textbf{31.36} \\ \hline

\hline 
\hline
\multicolumn{1}{|c|}{\multirow{13}{*}{HDA}} &  Dragonskin & \multicolumn{1}{c|}{86.65} & \multicolumn{1}{c|}{95.94} & \multicolumn{1}{c|}{49.22} & \multicolumn{1}{c|}{84.05} & \multicolumn{1}{c|}{75.22} & \multicolumn{1}{c|}{67.16} & \textbf{27.52}\\ 
\cline{2-9}  &    Ecoflex    & \multicolumn{1}{c|}{80.02} & \multicolumn{1}{c|}{95.58} & \multicolumn{1}{c|}{36.29} & \multicolumn{1}{c|}{51.94} & \multicolumn{1}{c|}{43.33} & \multicolumn{1}{c|}{46.94} & \textbf{15.61}\\ 
\cline{2-9} &  Gelafix     & \multicolumn{1}{c|}{85.32} & \multicolumn{1}{c|}{78.42} & \multicolumn{1}{c|}{59.87} & \multicolumn{1}{l|}{36.51} & \multicolumn{1}{c|}{41.22} & \multicolumn{1}{c|}{48.95} & \textbf{21.64} \\ 
\cline{2-9}   &  Gelatin  & \multicolumn{1}{c|}{74.42} & \multicolumn{1}{c|}{76.31} & \multicolumn{1}{c|}{37.74} & \multicolumn{1}{c|}{42.73} & \multicolumn{1}{c|}{29.39} & \multicolumn{1}{c|}{56.68} &\textbf{19.09}  \\ 
 \cline{2-9}  & Glue   & \multicolumn{1}{c|}{83.40} & \multicolumn{1}{c|}{94.33} & \multicolumn{1}{c|}{65.48} & \multicolumn{1}{c|}{67.90} & \multicolumn{1}{c|}{42.77} & \multicolumn{1}{c|}{45.66} &\textbf{29.93}  \\ 
\cline{2-9} &   Knetosil     & \multicolumn{1}{c|}{91.03} & \multicolumn{1}{c|}{92.38} & \multicolumn{1}{c|}{40.03} & \multicolumn{1}{c|}{23.20} & \multicolumn{1}{c|}{24.45} & \multicolumn{1}{c|}{29.81} & \textbf{7.48} \\ 
\cline{2-9}  &  Latex    & \multicolumn{1}{c|}{93.10} & \multicolumn{1}{c|}{96.94} & \multicolumn{1}{c|}{52.50} & \multicolumn{1}{c|}{25.30} & \multicolumn{1}{c|}{24.13} & \multicolumn{1}{c|}{29.96} & \textbf{9.19}\\
\cline{2-9} &   Modelling clay   & \multicolumn{1}{c|}{84.94} & \multicolumn{1}{c|}{92.17} & \multicolumn{1}{c|}{64.56} & \multicolumn{1}{c|}{21.46} & \multicolumn{1}{c|}{27.61} & \multicolumn{1}{c|}{24.33} &\textbf{1.63}  \\
\cline{2-9}    &  Mouldable glue  & \multicolumn{1}{c|}{85.26} & \multicolumn{1}{c|}{94.18} & \multicolumn{1}{c|}{54.67} & \multicolumn{1}{c|}{21.81} & \multicolumn{1}{c|}{27.82} & \multicolumn{1}{c|}{20.03} & \textbf{5.02}\\ 
 \cline{2-9}  &  Paper printout  & \multicolumn{1}{c|}{88.67} & \multicolumn{1}{c|}{86.55} & \multicolumn{1}{c|}{57.54} & \multicolumn{1}{c|}{65.86} & \multicolumn{1}{c|}{24.51} & \multicolumn{1}{c|}{71.03} & \textbf{0}\\ 
\cline{2-9} & Playdoh    & \multicolumn{1}{c|}{85.03} & \multicolumn{1}{c|}{93.01} & \multicolumn{1}{c|}{55.16} & \multicolumn{1}{c|}{58.12} & \multicolumn{1}{c|}{54.39} & \multicolumn{1}{c|}{68.78} & \textbf{20.96}\\ 
\cline{2-9}  &  Silly putty   & \multicolumn{1}{c|}{86.90} & \multicolumn{1}{c|}{86.21} & \multicolumn{1}{c|}{58.65} & \multicolumn{1}{c|}{66.87} & \multicolumn{1}{c|}{71.30} & \multicolumn{1}{c|}{47.58} & \textbf{21.75}\\
 \hline
\end{tabular}}
\end{table*}
The results of our experiments are presented in Tables~\ref{tab:EER compare table experiment1},~\ref{tab:EER compare table experiment2} and~\ref{tab:EER compare table experiment3} for Experiments 1--3 respectively. Across all three experiments, we observe the following:
\begin{itemize}  [leftmargin=*,noitemsep, topsep=0pt,parsep=0pt,partopsep=0pt]
\item Among all the unsupervised methods, our DDPM-based approach consistently achieves the best BPCER on the vast majority of PAIs.
It is only slightly worse than the ViT OC SVM on Playdoh PAI for Experiment 1.
In the remaining cases where DDPM is not the best, the BPCER for all methods are extremely high, so the ranking of different methods is a poor comparison.
\item The photopaper and printed attacks are considered to be some of the most challenging PAIs, and DDPM has substantially lower BPCER compared to other approaches.
\end{itemize}

From Table~\ref{tab:EER compare table experiment1}, we make the following observations regarding the results of Experiment 1:
\begin{itemize}  [leftmargin=*,noitemsep, topsep=0pt,parsep=0pt,partopsep=0pt]
    \item The DDPM trained on CLARKSON has a clear degradation of BPCER when tested against the out-of-distribution NTNU dataset.
    \item In contrast, the DDPM method beats every baseline model on the HDA dataset.
    \item This indicates that the model generalization may be affected by input data with different capture settings. In particular, the CLARKSON and HDA datasets could have similar capture settings, whereas the CLARKSON and NTNU datasets could have dissimilar settings.
\end{itemize}

From Table~\ref{tab:EER compare table experiment2}, we make the following additional observations regarding the results of Experiment 2:
\begin{itemize}  [leftmargin=*,noitemsep, topsep=0pt,parsep=0pt,partopsep=0pt]
    \item By training on the NTNU dataset only, DDPM performance is very clearly degraded for CLARKSON and HDA datasets. As explained above, this is to be expected.
    \item However, we observe that the three attacks on the NTNU dataset offer slightly better BPCER.
\end{itemize}

Finally, Table \ref{tab:EER compare table experiment3} allows us to draw the following conclusions about Experiment 3, where we train all models on the combined CLARKSON and NTNU datasets:
\begin{itemize}  [leftmargin=*,noitemsep, topsep=0pt,parsep=0pt,partopsep=0pt]
    \item Compared to Table~\ref{tab:EER compare table experiment1}, the additional training data leads to degraded performance for DDPM on CLARKSON PAIs, especially in the challenging photopaper setting. 
    \item However, we also obtained substantially better BPCER performance compared to Table~\ref{tab:EER compare table experiment2} on both NTNU and HDA datasets, so there is a tradeoff involved here in the choice of datasets.
    \item In the combined dataset setting, DDPM consistently offers the lowest BPCER out of all approaches.
    \item Overall, we observe that the proposed method can be unstable if the capture environments of the training and test samples are different.
    It may be helpful to include multiple capture environments in training data.
    Nevertheless, the generalization of the detection method under different cameras, light environments, and capture distances should be further considered in realistic deployments. 
\end{itemize}

\subsection{Choice of Image Similarity Metric}
Since our methodology in Section~\ref{sec:method} is compatible with any choice of image similarity metric, we conduct an alternative experiment with our DDPM model using different metric choices.
For simplicity, we only carry out this experiment on the CLARKSON dataset using four PAIs.
We experimented with the following metrics:

\begin{itemize} [leftmargin=*,noitemsep, topsep=0pt,parsep=0pt,partopsep=0pt]
    \item Mean-Square Error (MSE): the average squared difference among all pixels between the source image and the reconstructed target image.
    \item Structural Similarity Index Measure (SSIM)~\cite{wang2004image} which measures the structural (perceived) similarity between two images.
    \item Learned Perceptual Image Patch Similarity (LPIPS), which is our proposed similarity metric.
\end{itemize}

\begin{table}[h]
\centering
\caption{BPCER @ APCER=10\% of DDPM model against four different PAIs using various similarity metrics (best in \textbf{bold}).}
\label{tab:similarity metrics EER}
{\small\begin{tabular}{|c|c|c|c|}
\hline
PAI & MSE & SSIM &  LPIPS\\ \hline
Ecoflex & 14.10 & 14.89 & \textbf{6.34} \\ \hline
Photopaper & 19.02 & 29.99 & \textbf{12.89} \\ \hline
Playdoh & 17.18 & 20.23 & \textbf{9.46} \\ \hline
Woodglue & 12.02 & 13.82 & \textbf{3.64} \\ \hline
\end{tabular}}
\end{table}

The results are shown in Table \ref{tab:similarity metrics EER}.
We observe that the deep feature-based LPIPS metric achieves superior classification performance over MSE and SSIM across all four PAIs of the CLARKSON dataset.
This justifies our choice of using LPIPS as the default similarity metric.

\subsection{Fingerphoto Fidelity for Generative Models}
In this section, we present quantitative measurements of the reconstruction quality for our fingerphoto models using the Frechet Inception Distance (FID) score, which is a commonly used metric to assess the quality of image generation~\cite{DBLP:conf/nips/HeuselRUNH17}.
We again used the CLARKSON dataset for simplicity and calculated each model's FID score on the set of bona fide test samples and on each attack PAI test set.

The results in Table~\ref{tab:FID} are not fully conclusive regarding the relationship between FID and detection error rates, but we can draw some partial observations.
Firstly, DDPM achieves the best FID score for \emph{bona fide} fingerphoto reconstruction across all models.
This shows that DDPM may be a viable model for purposes other than fingerphoto PAD, particularly those that require high-quality outputs, e.g., fingerphoto de-occlusion.
Furthermore, it is interesting to observe that with DDPM, there is a clear gap (1 order of magnitude) between the FID for bona fide samples and for the respective PAIs.
However, FID scores are not the only predictors of PAD performance. For CAE and VAE, the results are surprisingly inverted, i.e., bona fide samples have worse FID scores than attack samples.
This may indicate that quality of CAE and VAE fingerphoto generation are too poor for FID scores to be meaningful.

\begin{table}[]
\caption{FID score for different image reconstruction techniques.}
\label{tab:FID}
\centering
{\small
\begin{tabular}{|c|c|c|c|c|c|}
\hline
 Image type& CAE & VAE & SGA &DDPM  \\ \hline
  Bona fide& 438.28 & 475.97 & 54.30 & 8.84 \\ \hline
  PAI: Ecoflex& 419.73 & 423.25 & 103.53 & 107.94 \\ \hline
  PAI: Photopaper& 348.33 & 391.96 & 84.65 & 74.17 \\ \hline
  PAI: Playdoh& 247.74 & 343.37 & 184.95 & 228.33 \\ \hline
PAI: Woodglue& 427.49 & 444.36 & 105.93 & 166.50 \\ \hline
\end{tabular}}
\end{table}

\subsection{Discussion of Misclassified Samples}
\begin{figure}[t!]
\begin{center}
\includegraphics[width=0.8\linewidth]{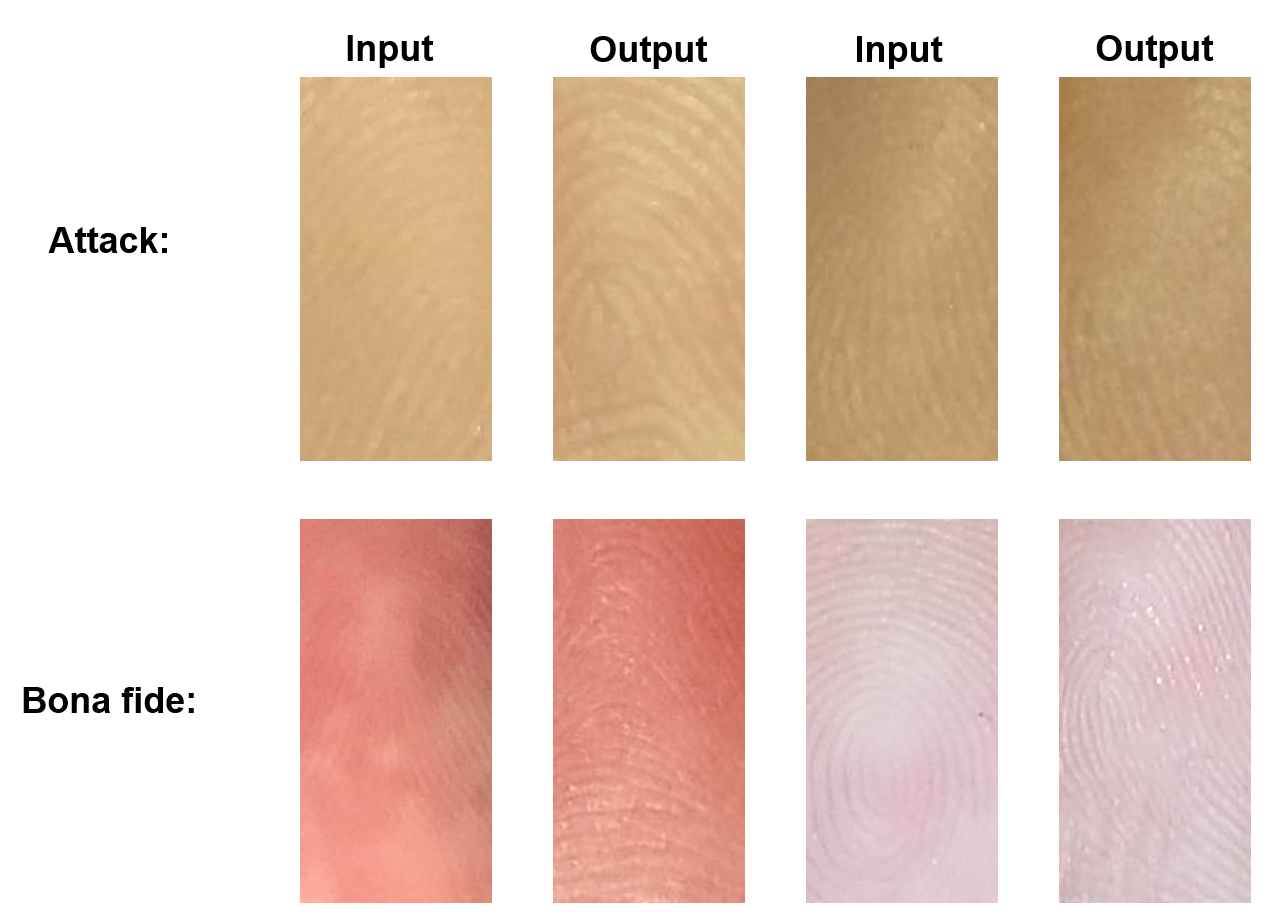}
\end{center}
   \caption{Misclassification cases for DDPM input/output pairs. The first row shows two attack examples and the second row shows two bona fide examples.}
\label{fig:mis_case}
\end{figure}

In this section, we provide insights into the misclassification of samples using the proposed fingerphoto PAD method.
Four misclassified pairs of bona fide and attack samples are illustrated in Figure~\ref{fig:mis_case}.

The first row shows the case in which attack samples are misclassified as bona fide inputs, i.e., the LPIPS similarity between the attack image input and reconstruction is high.
One observation in this case is that the texture is not clear in the input sample, which leads to an issue in which the output sample is also at low resolution; the similarity measurement fails to distinguish these two low resolution samples.

Similarly, the bona fide misclassification pairs in the bottom row also suffer from low-quality inputs. The input samples were unclear because of light reflection. However, DDPM reconstructs the sample to a higher resolution with a clear texture. Hence, the distance between the input and output increases according to the similarity metric.
Based on these observations, we hypothesized that a key challenge for DDPM-based fingerphoto PAD is the capture quality of the input samples. In future work, we will further investigate this hypothesis.

\section{Conclusion}
\label{sec:conclusion}

Fingerphoto presentation attacks have been frequently researched and demonstrated in recent years.
This study focuses on tackling the generalization and scalability challenges in fingerphoto presentation attack detection (PAD) arising from the need to rapidly adapt to new types of presentation attack instruments (PAIs).
We propose a novel combination of two deep learning methods, DDPM and LPIPS, for unsupervised fingerphoto PAD.
Our approach produces realistic fingerphoto reconstructions and yields very promising results compared to baseline approaches. 

We are also investigating the potential dual applications of the trained DDPM model going beyond PAD.
For example, after using the reconstructed fingerphoto for PAD, it may also be used as part of a preprocessing pipeline to improve the captured image quality of fingerphotos or for the de-occlusion of partial fingerphoto images.

\paragraph{Acknowledgements.} This work is carried out under the OFFPAD project funded by the Research Council of Norway (Project No. 321619).
This work is also supported by the Digital Trust Centre (DTC) Research Grant funding: DTC-RGC-01, and by the National Research Foundation, Prime Minister’s Office, Singapore, and the Ministry of Communications and Information, under its Online Trust and Safety (OTS) Research Programme (MCI-OTS-001). Any opinions, findings, conclusions, or recommendations expressed in this material are those of the author(s) and do not reflect the views of the National Research Foundation, Prime Minister’s Office, Singapore, or the Ministry of Communications and Information.
Hailin Li is supported by the A$^*$STAR Research Attachment Programme (ARAP) at I$^2$R.

\bibliographystyle{ieee}
\bibliography{egbib}

\end{document}